# Multi-Criteria Comparison as a Method of Advancing Knowledge-Guided Machine Learning

**Jason L. Harman[1] & Jaelle Scheuerman[2]**
1 Louisiana State University
2 Naval Research Laboratory

### Abstract

This paper describes a generalizable model evaluation method that can be adapted to evaluate AI/ML models across multiple criteria including core scientific principles and more practical outcomes. Emerging from prediction competitions in Psychology and Decision Science, the method evaluates a group of candidate models of varying type and structure across multiple scientific, theoretic, and practical criteria. Ordinal ranking of criteria scores are evaluated using voting rules from the field of computational social choice and allow the comparison of divergent measures and types of models in a holistic evaluation. Additional advantages and applications are discussed.

## Introduction

Modern advances in Artificial Intelligence (AI) and Machine Learning (ML) have led to AI/ML implementations across numerous public and private contexts. With the proliferation of AI/ML decision models, an increasing number of scholars have pointed out the shortcomings of ML when compared to scientifically informed theories and prediction models such as lack of generalizability and adverse impact. The core of these arguments are that ML models can be a-theoretic black boxes, meaning they produce predictions based only on given data to optimize the accuracy of some output with no theoretical rationale for predictions and no explainable process. Beyond these major issues inherent to ML, there is additional evidence that advances in accuracy of decision making from ML are overstated if not absent when compared to simpler cognitively inspired decision models that are both explainable, theoreticaly consistent, and less amenable to adverse impact (*e.g.* Harman & Scheureman, 2022; Martignon, Katsikopoulos, & Woike, 2008; Frederik & Martijn, 2019; Gigerenzer & Gaissmaier, 2011; Katsikopoulos, 2011, Katsikopoulos et al. 2018). In this paper we advance a multi-criteria evaluation and comparison method recently introduced in the field of decision making. This method allows for the evaluation of ML models across multiple theoretic and scientific criteria. The method is also flexible and adaptable to multiple fields and contexts and can both incentivize better models and evaluate extant models.

## Background

The current work stemmed from critiques of a modeling competition in Psychology (Harman et al., 2021). The Choice Prediction Competition (CPC; Erev, Ert, & Plonski, 2017) was a prediction competition designed to promote generalizable prediction models for human decision making. The problem being addressed was that decision making models in Psychology and Cognitive Science tend to be built to explain choice anomalies (patterns of human choice that violate the rational axioms of expected utility). With a growing multitude of these anomalies (some of which contradict each other), the number of decision-making models created to account for them has also grown with few models designed to account for decision making across diverse contexts.

Erev et al., created a unique paradigm that could replicate multiple different decision-making paradigms and replicate 14 well known choice anomalies which had yet to be accounted for by a single model. They then invited research groups to enter model that accounted for all known anomalies and predicted new data the best. 25 models were able to enter the competition (by passing the threshold of accounting for all 14 historic anomalies). Of the modles entered, 14 were variants of a baseline model the organizers provided as an example, 6 were variants of Prospect Theory (Khaneman & Tversky, 1979), 4 were machine learning models, and 1 was a cognitive process model based on Instance Based Learning (Gonzalez, Lerch, & Lebiere, 2003). Of note, the

4 ML models fit the calibration data well, but when predicting new data were far and away the worst models in the competition. All of the leading models were variants of the baseline provided and the PT variants along with the process model finished in the middle of the pack.

Noting that there was limited variety in the types of models entered and that only one model hand important theoretical and scientific qualities (e.g. identifiable process assumptions, parsimony), Harman et al. (2021) outlined multiple factors that limited the impact such a competition could have and provided a possible solution. The main theme of many of the critiques was the reliance on a single evaluative criterion; minimized prediction error (MSD in this case). Harman et al. outline how using a single evaluative criterion limits the type and variety of models entered (and subsequent insights from comparing different types of models) by incentivizing predictive accuracy only. While predictive accuracy is an important aspect of a predictive model, it is not the only important aspect of a model. Generalizability, explainability, parsimony, and falsifiability are a few of the other qualities that are desirable for a good model. TO provide a solution to these shortcomings, Harman et al., introduced a method of quantifying and combining additional desirable criteria (e.g. generalizablity, explainability, adverse impact) into a method of evaluating models across multiple criteria. The initial work was formalized for scientific competitions of decision-making models but can be readily applied to any AI/ML or predictive modeling evaluation.

## Mulit-Criteria Model Evaluation

Harman et al. outline multiple reasons why evaluating predictive decision models on a single evaluative criterion (e.g. predictive accuracy) have disadvantages that incentivize problematic characteristics of models (e.g. lack of generalizability, overfitting, lack of explainability) and disincentivize desirable characteristics of models. Their unique solution was to design an evaluation system (originally to be used in modeling/prediction competitions) which evaluates models along multiple criteria at once, adding unique and emergent insights into model comparisons. The first prerequisite for a multi-criteria comparison or competition is establishing a taxonomy of desirable characteristics of a model. Following is the taxonomy Harman et al. developed for modeling competitions in human decision making;

1. *Theoretical criteria*
1.1 Intuitive understanding- A model should be able to guide intuitive predictions and interventions/prescriptives in the real world. (see Katsikopoulos, 2020; 2014)
1.2 Broad scope- a model should be able to be applied to (or easily adapted to) various scenarios / paradigms. (see Busemeyer & Wang, 2000)

2. *Psychological Criteria*
2.1 Realistic knowledge- Predicted behavior should not be based on information participants are not likely to have, or is hard to obtain. (Meir, Lev, & Rosenschein, 2014)
2.2 Realistic capabilities- Predicted behavior should not rely on complex computations, non-trivial probabilistic reasoning, etc. (Busemeyer & Diederich, 2009)
2.3 Identifiable process assumptions- A model should rely on identifiable and testable psychological processes. (Weber & Johnson, 2012)

3. *Scientific Criteria*
3.1 Parsimony- a model should have as few parameters as possible, and parameters should be meaningful. (Kuhn, 1977)
3.2 Predictive power / validation - A model should be able to predict new behavioral data with accuracy. (Busemeyer & Wang, 2000)
3.3 Reproductive Power - a model should be able to reproduce common phenomena. (Erev et al., 2017)
3.4 Testability / Falsifiability- A model should produce predictions that could be falsified, or predict behavior that would not happen. (Popper, 1934/1959; Roberts & Pashler, 2000)

The taxonomy detailed by Harman, et al. represents the major desirable characteristics of a cognitive decision-making model. One of the advantages to the method developed by Harman et al. is that it can be adapted as needed by different fields. Models that are not psychological in nature for example may exclude category 2 all together while adding additional criteria. Likewise, additional criteria could be added specific to different fields and goals. For example, a more specific taxonomy for explainable AI (XAI) could include criteria such as:

4. *Explainability Criteria*
4.1 Common Explainability - a human user should be able to generate an adequate mental model of the AI decision process.
4.2 Formal Explainability/interpretability - A model's decision should be traceable or reproducible.
4.3 Trust – A model should produce predictions that are trusted by human users and meet their expectations.
5. *Ethical Criteria*
5.1 Adverse Impact – A model should not produce differential error rates correlated with race, gender, income, or other population characteristics.

These are very general ideas for additional criteria relevant to XAI, but they serve to illustrate the flexibility of the evaluation method.

## Quantifying Criteria

The key to the multi-criteria evaluation procedure is that all criteria be quantified at least ordinally (including dichotomous rankings). Harman et al. detail multiple ways that their outlined criteria could be quantified. Predictive power is a straightforward quantification of minimized prediction error using measures such as MSD. Other criteria, such as intuitive understanding, broad scope, or falsifiability are more flexible. At the simplest level, quantifying some of these criteria could be done in a competition by model builders checking a box, the model is/is not falsifiable. Alternatively, competition organizers could appoint independent judges to provide those ratings. A more in depth measure of something like broad scope could provide several scenarios/paradigms for a model to predict and produce a count of how many paradigms the model can be generalized to.

The key to this step is that each criterion is assigned a rank of some sort. Harman et al., discuss in depth how competition organizers have flexibility in doing this and how competition goals could be reflected in the quantification mechanisms. As will be seen in the next section, a continuous measure will have a larger impact on models' final evaluations. As an example, consider a modeling competition concerned with selecting employees from a large pool with multiple pre-employment measures. So, if an organizer is primarily interested in whether a selection algorithm produces adverse impact for example – a measure such as, the difference in proportion of minority /women candidates of the selection pool and the chosen people would be quantified continuously. If the organizers were primarily interested in predicting the best performers, this measure of adverse impact could be dichotomous with a threshold (i.e. if the difference in proportion of minorities is less than X, the model scores 1 else 2). A middle ground could also be established where multiple bins are created for the adverse impact score representing a categorical measure; 1(0-1%), 2(1-3%), 3(3-5%), etc.

What's important in the examples above is that each important criterion is quantified. Though organizers may minimize the importance of a criteria through its quantification, the fact that it is still measured has multiple important consequences. To name a few; models (and model builders) are incentivized to consider different criteria, a competition is opened to a larger variety of model types, and importantly post hoc comparisons of models are enriched by clearly showing a models' standing relative to other models across a variety of features. The major quantitative advancement proposed by Harman et al., was the adoption of voting rules from the field of computational social choice to perform direct model comparisons across multiple criteria at once.

## Evaluating Models

To evaluate candidate models (and select a winner for modeling competitions) Harman et al. propose a combination of Condorcet and Borda rule voting where models are ranked ordinally on each criteria. If one model is a Condorcet winner (better than every other model on a majority of criteria) the competition is over (see Fishburn, & Gehrlein, 1977 for a detailed discussion of Condorcet consistency), and if there is no Condorcet winner a Borda voting rule is applied, where models are given points based on their rank on each criteria, with agreed upon tie breakers in the cases of Borda ties.

As an example consider hypothetical results from two simplified competitions (Figure 1). In both competitions, the first criterion is an ordinal ranking with no ties such as MSD. The second, fourth, and fifth criteria are binary criteria with a model that satisfies the criteria ranked 1 and models that fail to satisfy the criteria ranked 2. Criterion 3 represents an ordinal ranking with ties, such as accounting for historical phenomena where a model could account for all phenomena, all but one, all but two etc.

**Figure 1**

*Hypothetical competition rankings for two modeling competitions*

Competition 1

|         | C1 | C2 | C3 | C4 | C5 |
|---------|----|----|----|----|----|
| Model 1 | 3  | 1  | 2  | 2  | 1  |
| Model 2 | 1  | 1  | 1  | 1  | 1  |
| Model 3 | 2  | 1  | 1  | 1  | 2  |
| Model 4 | 4  | 2  | 1  | 1  | 2  |
| Model 5 | 5  | 1  | 3  | 1  | 2  |

Competition 2

|         | C1 | C2 | C3 | C4 | C5 |
|---------|----|----|----|----|----|
| Model 1 | 3  | 1  | 2  | 2  | 1  |
| Model 2 | 1  | 2  | 1  | 1  | 1  |
| Model 3 | 2  | 1  | 1  | 1  | 1  |
| Model 4 | 4  | 2  | 1  | 1  | 2  |
| Model 5 | 5  | 1  | 3  | 1  | 2  |

*Note.* Figure 1 shows hypothetical rankings of 5 models across 5 criteria (C1-C5).

To first establish whether a Condorcet winner is present, all pairwise comparisons are performed with a model that is ranked above another model in a majority of criteria being superior. For example, in the first competition Model 3 is superior to Model 1 as it ranks higher than Model 1 on three out of five criteria. These pairwise comparisons can be illustrated using an edge graph (Figure 2) where a model that is superior to another has a line pointing away from it to the dominated model. A tie between models would be represented with a double headed arrow. A Condorcet winner then would have all possible lines pointing away from it. Examining Figure 2 it is clear that Model 2 is a Condorcet winner in competition 1 and would be declared the winner with no further computation.

**Figure 2**
*Edge graphs for competitions 1 and 2*

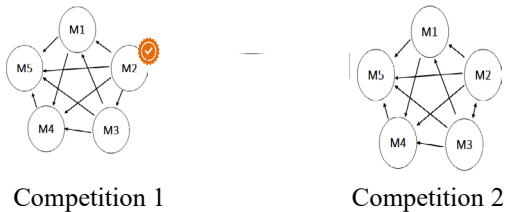

Competition 1          Competition 2

*Note*. Figure 2 displays an edge graph of the 5 hypothetical models from Figure 1. Directional arrows represent a model that dominates another model and double headed arrows represent a tie.

Competition 2 does not have a Condorcet winner as Models 2 and 3 are tied (each beats the other on one criterion and they are tied on the remaining three criteria). In this case a Borda run-off would be performed. In Borda rule voting, models are assigned points to their rank on each criterion with more points for higher ranks. Because Criteria 1 and 3 have more than two ranks, winners of these criteria would receive an advantage. Figure 3 shows the Borda count for each model in competition 2. In many cases a Borda run-off would determine a winner when no Condorcet winner was present. In this example however, models 2 and 3 remain tied after the Borda run off. There are two possibilities in this case; one is that organizers could agree that ties are acceptable, and two models would be declared winners; the second alternative would be using an ordinal criterion that the organizers believe to be the most important to declare a final winner. In the case of the CPC and many other competitions this would be MSD.

Note that in these two fictitious examples, the model that minimized MSD more than all others is still declared the winner. The process of getting there though, opens the door for more diverse models in the competition and more methods for comparing model performance and testing auxiliary hypotheses, multiplying the potential insights that could be gained from a single competition. Additionally, the relative importance of specific criteria (i.e., prediction) could still be determined by competition organizers via binary vs. rank ordering. In the CPC for example, all of the models that qualified would be ranked 1 on a reproductive power criterion, making the strictly ordinal prediction criterion more discriminating. Not only would a multi criteria competition set up improve the diversity of models entered, this more in depth model comparison procedure could clarify the best properties of the ultimate winner. In the final results of the CPC, 12 of the top models were statistically indistinguishable (Erev, et al., 2017, p. 389) and the winner was basically a random draw. With multiple criteria, further comparisons have the possibility of distinguishing competing models beyond their statistical tie. A more detailed outline of setting up and running a multi-criteria competition is provided in Harman et al., 2021 supplemental online material.

In addition to allowing direct model evaluation across multiple differing criteria, this structure also provides more insight into relative model performance. For example, in a traditional competition a model may win because its error term is slightly lower than other models with no other insights gained. In a multi-criteria competition that model may win because it has a slightly lower error term and has more evenly distributed errors between gender and race than other models etc. Key for the current topic, the multi-criteria method would allow the promotion of scientifically inspire qualities in the creation of ML models.

## Discussion

The current paper introduces a unique method for comparing and evaluating ML models across multiple empirical and theoretical model features. While the current work does not directly address how to improve knowledge guided ML, it provides a way to quantify and evaluate ML models in their implementation of scientific knowledge, a key step in advancing KGLM. Some of the advantages of this approach include the flexibility of establishing and quantifying the features most desirable in a good model and the ability to posteriorly decompose models' performance in a competition to directly explain why one model may be preferred to another. Multiple tertiary benefits include opening modeling competitions to a variety of modeling approaches including simple approaches that may outperform ML.

As an example of this final point, the authors entered a recent ML competition with a simple model inspired by the selection procedure presented here (Harman & Scheuerman, 2022). The 3rd annual SIOP Machine Learning competition was designed to promote ML algorithms for personnel selection that both predicted successful hires while minimizing adverse impact. Using a large rea world data set from Walmart that included pre-hire demographics and employment screenings, competition participants were tasked with creating a prediction model that would select the top 50% of employees (retention and performance after hire) while maintaining the proportion of minority and female applicants in the initial pool. Teams were able to train and test models on a public data set and leaderboard and were evaluated with a private dataset/leaderboard. The DV in this case was a combined metric of successful prediction – adverse impact.

To test the robustness of the selection methodology resented here, we entered a simple (non-ML) model with 3 rules: rank order candidates on each variable, perform a Borda count on those rankings, and select the top 50% of Borda scores. Over 60 models were entered into the final competition and our simple set of rule finished 9th, far outperforming a majority of the ML models entered and being comparable to the top performing models. In addition to the prediction accuracy of this simple method, it retained positive qualities of the method presented here, namely that it was simple to implement and was completely explainable (e.g. a candidate not selected could see their relative performance on each variable and understand why the were beaten out).

With recent advances such as those in pareto optimization, and a growing consensus on some of the shortcomings of current ML, the future of KGML is promising. As advances increase, there is a need to have a standard method of gauging those advances and avoid overselling results. We believe that the multi-criteria comparison procedure presented here may be part of that solution.